%% file: lrec-coling2024.tex
\documentclass[10pt, a4paper]{article}

\usepackage[]{lrec-coling2024} 
\usepackage{latexsym}
\usepackage{graphicx}
\usepackage{subcaption}

\usepackage{amsmath}
\usepackage{multirow}

\title{ECtHR-PCR: A Dataset for Precedent Understanding and Prior Case Retrieval in the European Court of Human Rights}

\name{Santosh T.Y.S.S,  Rashid Gustav Haddad, Matthias Grabmair} 

\address{School of Computation, Information, and Technology; \\ Technical University of Munich, Germany
\\
\{santosh.tokala, rashid.haddad, matthias.grabmair\}@tum.de\\}

\abstract{
In common law jurisdictions, legal practitioners rely on precedents to construct arguments, in line with the doctrine of \emph{stare decisis}. As the number of cases grow over the years, prior case retrieval (PCR) has garnered significant attention. Besides lacking real-world scale, existing PCR datasets do not simulate a realistic setting, because their queries use complete case documents while only masking references to prior cases. The query is thereby exposed to legal reasoning not yet available when constructing an argument for an undecided case as well as spurious patterns left behind by citation masks, potentially short-circuiting a comprehensive understanding of case facts and legal principles. To address these limitations, we introduce a PCR dataset based on judgements from the European Court of Human Rights (ECtHR), which explicitly separate facts from arguments and exhibit precedential practices, aiding us to develop this PCR dataset to foster systems' comprehensive understanding. We benchmark different lexical and dense retrieval approaches with various negative sampling strategies, adapting them to deal with long text sequences using hierarchical variants. We found that difficulty-based negative sampling strategies were not effective for the PCR task, highlighting the need for investigation into domain-specific difficulty criteria. Furthermore, we observe performance of the dense models degrade with time and calls for further research into temporal adaptation of retrieval models. Additionally, we assess the influence of different views , Halsbury's and Goodhart's, in practice in ECtHR jurisdiction using PCR task.
 \\ \newline  }

\begin{document}

\maketitleabstract

\section{Introduction}
\input{text/introduction}

\section{Related Work}
\input{text/related}

\section{ECtHR-PCR Dataset}
\input{text/dataset}

\section{Models}
\input{text/models}

\section{Experiments}

\input{text/experiments}

\section{Conclusion}
\input{text/conclusion}

\section{Limitations} 
\input{text/limitations}

\section{Ethics Statement}
\input{text/ethics}


\section{Bibliographical References}\label{sec:reference}

\bibliographystyle{lrec-coling2024-natbib}
\bibliography{lrec-coling2024}


\end{document}

%% file: text/introduction.tex
Legal systems globally can be broadly categorized into two main frameworks: common law and civil law \cite{joutsen201942}. In common law countries, such as the United States, England and India, published judicial opinions, known as case law, hold primary significance. Conversely, civil law systems, as observed in countries like China, Japan, France and Germany, place greater emphasis on codified statutes. Naturally, these distinctions are not always rigidly defined and in reality, many countries have adopted a combination of features from both.

Legal practitioners in common law jurisdictions rely on existing case decisions, known as precedents, as a vital source of law, 
based on the doctrine of \emph{stare decisis}, which can be translated from Latin as "to stand by the decided cases." It emphasizes that when a case being considered shares similarities with past cases, it should be treated in a manner consistent with those precedents \cite{duxbury2008distinguishing,lamond2005precedents}. By examining previous rulings and citing relevant precedents, legal professionals establish the relevant law applicable to the current case and strengthen their arguments \cite{mandal2017overview,shulayeva2017recognizing}. 

With the increasing volume of cases, there is a growing demand for automatic precedent retrieval systems to aid practitioners by providing prior cases relevant to the current case. The task of Prior Case Retrieval (PCR) has gained attention in the legal and information retrieval communities, leading to the curation of datasets such as  COLIEE \cite{kim2018coliee} and IRLeD \cite{mandal2017overview}, which have facilitated advancements in PCR using case law from the Federal Court of Canada and Indian Supreme Court respectively. These datasets define relevance criteria as the precedents cited in the query document, utilizing citation as a signal to construct the dataset. While civil law systems may not directly involve prior cases in the final judgment, recognizing their importance as references for consistency 
in decision-making, efforts have been made to curate PCR datasets tailored to the Chinese law system, such as CAIL-SCM \cite{xiao2019cail2019} and LeCaRD \cite{ma2021lecard}, which relied on legal professionals' expertise due to the absence of a citation structure in civil law case documents, resulting in smaller dataset sizes.



Prior PCR datasets in common law (COLIEE, IRLeD) treated entire case documents as queries, suppressing citations to prior cases by using "FRAGMENT SUPPRESSED" / "CITATION" tags. However, this approach fails to effectively address the problem and raises concerns about evaluating PCR systems effectiveness. Case documents comprise both factual aspects, describing the circumstances of the case, and argumentative aspects, encompassing judges' reasoning, citations to prior cases, legal tests, analysis, and explanations for their decisions. By solely suppressing citations, models or systems tend to exploit the pattern of citation suppression fragments and rely exclusively on the citance (text preceding the citation) to identify relevant prior cases. 
Additionally, the problem is exacerbated when dealing with extractive (verbatim) citances, where the citance in the query  is directly extracted from the cited document \cite{rossi2021verbcl}. This simplifies the PCR task to an exact matching problem, potentially overestimating the system's performance, disregarding the nuanced aspects of the case. 


Moreover, in a practical scenario, the arguments section of a case is often available only after the final verdict has been delivered, while only the factual aspects are accessible prior to the verdict. Thus, the system must rely solely on understanding and analyzing the facts to determine the relevance of prior cases in reaching the outcome of the current case.  However, identifying relevance itself presents a challenge, as legal scholars may not unanimously agree on what factors constitute the \emph{ratio decidendi}, which refers to the binding reasons for a decision that have an impact on subsequent cases \cite{valvoda2021precedent}. While \citealt{halsbury1815laws} contends that the judge's reasoning and arguments are what bind, \citealt{goodhart1930determining} argues for the analogy between the facts of the precedent and the current case. 
Hence, the PCR model should ideally learn the nature of relevancy based on the distribution of available data and account for the divergent perspectives on what factors contribute to the ratio decidendi.

Given the limitations of existing datasets, it is crucial to develop a PCR dataset that enables systems to acquire a comprehensive understanding of the case facts, legal principles, and the broader context to ensure the effective application of precedent in legal decision-making. In this work, we construct a dataset for the task of PCR within the context of European Court of Human Rights (ECtHR) \footnote{Our PCR dataset is available at \url{https://github.com/TUMLegalTech/ECHR-PCR}}, which adjudicates complaints by individuals against states about alleged violations of their rights as enshrined in the European Convention of Human Rights. The choice of ECtHR judgements is motivated by two primary reasons. Firstly, unlike many other courts, such as those in India and Canada, the ECtHR's case law documents explicitly separate the facts from the arguments, which is vital for ensuring that the queries used in the PCR system do not contain the argument/reasoning sections. Secondly, the ECtHR's citation practices in its arguments section indicate a reliance on precedential law, similar to common law countries. Although the ECtHR is an international court without a formal doctrine of stare decisis \cite{jacob2014precedents}, there is strong evidence suggesting a precedential nature. This evidence can be found in the court's own guidelines  \footnote{\url{www.echr.coe.int/Documents/Guide_ECHR_lawyers_ENG.pdf} and  \url{www.echr.coe.int/documents/50questions_eng.pdf}
} as well as in the works of former judges \cite{zupancic2016context} and legal scholars \cite{lupu2010role}. Through its extensive case-law, the ECtHR has played a vital role in shaping the interpretation of the convention, effectively transforming it into a living instrument. 

Using our curated ECtHR-PCR dataset, we benchmark both lexical and dense retrieval based approaches employing different negative sampling strategies, modifying them to hierarchical variants, to accommodate the typically long text sequences. We observe that difficulty-based negative sampling strategies, which proved effective in IR tasks, did not translate to  PCR task, highlighting the need to design domain-specific difficulty criterion. Further, we observe that the performance of dense models  degrade over time due to their inability to accommodate to newer evolving unseen documents, highlighting the need for temporal adaptation of retrieval models. 
We also put Halsbury’s and Goodhart’s views to the test in practice, by comparing retrieval performance using the facts or the reasoning of the documents alone.  
Our experimental evidence points out that Halsbury’s view is more widely practiced in ECtHR jurisdiciton, compared to Goodhart's. 

%% file: text/related.tex
\subsection{Existing PCR Datasets}
The field of legal information retrieval has seen several notable competitions and datasets for PCR. Information Retrieval from Legal Documents (IRLeD) \cite{mandal2017overview} uses 200 Indian Supreme Court cases as queries, with 1000 relevant prior cases (randomly sampling 5 per query) and 1000 irrelevant documents. 
One of the tasks in Competition on Legal Information Extraction/Entailment (COLIEE) \cite{kim2018coliee}, which has been held annually since 2014, 
focuses on PCR in Canadian case law. It includes 898 and 300 query cases as well as 4415 and 1564 candidates cases for training and testing test respectively. 
IRLeD and COLIEE datasets use citations as signals to identify relevant documents while citation markers are suppressed using special tags in the document content.

Additionally, there are other datasets for the civil law framework, specifically based on the Chinese legal system, that utilize expert annotations. The CAIL2019-SCM dataset \cite{xiao2019cail2019} comprises 8,964 triplets, each consisting of a query case, a relevant case, and an irrelevant case. Similarly, the LeCARD dataset \cite{ma2021lecard} includes 107 query cases and 10,700 candidate cases.

Compared to existing datasets, our dataset has notable differences. We only utilize the facts section of a case as the query, which reflects the practical scenario where the arguments section is typically inaccessible until after the final verdict. In contrast to previous datasets that randomly select a small number of irrelevant candidates, our dataset encompasses the complete case law of the European Court of Human Rights since 1960, providing a larger pool of potential candidate documents. This presents a greater challenge in identifying the relevant documents for each query. For more detailed distinctions between our dataset and existing ones, please refer to Table \ref{datasets_prior}.

\subsection{Tasks on ECtHR Corpora}
Previous works focused on the judgements corpus of the European Court of Human Rights (ECtHR) have primarily been limited to the task of Judgment Prediction, \cite{aletras2016predicting,medvedeva2020using,chalkidis2019neural,santosh2023leveraging,santosh2023Zero}, defined as predicting whether any article of the convention has been violated given a fact statement, argument mining \cite{habernal2023mining,poudyal2019using,poudyal2020echr}, vulnerability detection \cite{xu2023vechr}, and event extraction \cite{filtz2020events,navas2022whenthefact}. \citealt{valvoda2021precedent} has explored the two jurisprudential views of Halsbury and Goodhart on what constitutes a ratio based on the ECtHR corpus, modelling the question as a case outcome classification task.
To the best of our knowledge, our work is the first to study PCR on ECtHR jurisdiction. We offer this dataset to the research community to facilitate further advancements and research in the area of prior case retrieval.

%% file: text/dataset.tex

\subsection{Dataset Construction}
We create our ECtHR-PCR dataset in four steps: (i) Collecting case documents and filtering (ii) Parsing case documents into the facts and the reasoning (law) sections (iii) Extracting citations from each document (iv) Mapping the citations to the actual documents.
\newline

\noindent \textbf{Document Collection \& Filtering:} We gather the complete collection of documents as an HTML data dump from HUDOC\footnote{\url{https://hudoc.echr.coe.int/}}, the public database of the ECtHR, including their metadata. We apply a filtering process to retain only the English judgment documents based on the metadata ("Document Type: HEJUD"). Each judgment document in our dataset is associated with an application number, which serves as a unique identifier for individual applications submitted to the ECtHR.

The court may merge multiple applications for procedural purposes if they are related. Consequently, a single judgment document can be associated with multiple application numbers. Furthermore, there are instances where a case initially decided by a Chamber is later referred to the Grand Chamber upon the parties' request. In such scenarios, the Grand Chamber produces a new judgment document that retains the same application numbers. To mitigate potential conflicts in the citation mapping process during subsequent stages, we employ a strategy of creating distinct versions of these cases by appending a version number to each document, ensuring that they remain distinguishable by their application numbers.
\newline

\noindent \textbf{Parsing Documents:}
The judgement documents follow a specific structure, covering different aspects of the case, as outlined in Rule 74 of the Rules of the Court\footnote{\url{https://www.echr.coe.int/Documents/Rules_Court_ENG.pdf}} under different sections. Here is a concise overview of the primary sections: \emph{Procedure:} outlines the procedural steps, from the submission of the individual application to the final delivery of the judgment. \emph{The Facts:} encompasses the factual background of the case along with the procedure typically followed in domestic courts prior to the application being lodged with the Court under the sub-section \emph{The Circumstances of the Case} and also contains legal provisions of domestic law relevant to the case under the sub-section \emph{Relevant Law}\footnote{These legal provisions do not contain information about the articles of the ECtHR Convention. Rather they are provisions of domestic law and other pertinent international or European treaties and materials.}. \emph{The Law:} provides the legal reasoning to justify the specific outcome reached with respect to allegation of articles of the ECtHR convention.  It places its reasoning within a broader framework of established rules, principles, and doctrines derived from its previous case-law, and grounds the decision by providing an explicit reference to those judgement documents. \emph{Conclusion:} declares the Court's verdict regarding whether a violation of the alleged article has occurred or not.

Parsing the judgment documents into these sections presented challenges because the HTML lacked a consistent structure. To address this, we developed various hand-crafted rules to identify the headers of the sections. Additionally, we utilized regular expressions to classify the main section each header belongs to, taking into account inconsistencies in their naming patterns. As a result, we were able to extract the facts of the case from the \emph{The Circumstances of the Case} subsection and the reasoning sections from the \emph{Relevant Law} and \emph{The Law} sections. For a small number of documents (0.1\%) where the fact or reasoning sections were not obtained through parsing, we manually labeled them.
\newline

\noindent \textbf{Citation extraction:} We observe that citations to prior judgements by the court, referred to as "Strasbourg Case Law" (SCL), are available in the HUDOC case metadata. It appears that the SCL information has been manually compiled and often copied verbatim from the judgment text as the citation string. However, we discovered that ~60\% of the cases are missing SCL metadata, and even when SCL is present, it is incomplete. To address this issue, we developed three strategies for automatically extracting citations from the raw text of documents. Firstly, we employed regular expressions to identify the presence of application numbers referenced in the document. Secondly, we iteratively created several regular expressions to capture instances where the document mentioned 'v.' followed by variations of ECtHR country names and a specific date format (e.g., Schenk v. Switzerland, 12 July 1988, § 46, Series A no. 140). Lastly, we noticed that cases prior to 1999 had a different citation format (eg., Van Leuven and De Meyere judgment of 23 June 1981, Series A no. 43, pp. 25-26, para. 5) Therefore, we utilized the citation strings obtained from the SCL metadata and searched for exact matches within the documents to extract the relevant citations.
\newline

\noindent \textbf{Mapping Citations to Documents:}
The citation strings we obtained lack a consistent structure; however, they commonly include elements such as the plaintiff, defendant (optional), application number (optional) and date (optionally: day, month, and year). To ensure a high recall in extracting citations, we developed multiple heuristics for matching these citation patterns back to case documents. Whenever multiple versions of applications are available, we accurately point the citation to the exact version based on the date, ensuring that it points to the latest version that occurred prior to the document from which the citation originated.

\subsection{Manual Quality Inspection}
We evaluate the quality of our dataset using a random sample of 120 documents for which we manually extract all the citations and filter out the ones for which we could not resolve the outgoing citations, such as if the cited document was not in English or not available in the HUDOC database. Precision and recall of the automatically extracted citations was calculated per document and average values are displayed in Table \ref{manual}. 

Using SCL metadata alone results in low recall, as only 56 out of 120 documents have SCL metadata. Mapping citation strings to documents proves to be challenging even with SCL information, as reflected by the precision not being 1.0. Extracting application numbers mentioned in documents improves recall but suggests that not all citation formats include application numbers. Despite obtaining application numbers, precision is lower as some numbers resemble the regular expression of ECtHR application numbers but refer to external national case documents. Our method, combining SCL, application numbers, and citation string extraction using patterns like 'v.' and others, shows improved recall, indicating the high quality of our constructed dataset.


\begin{table}[]
\centering
\begin{tabular}{|l|c|c|}
\hline
              & \textbf{Recall} & \textbf{Precision} \\ \hline
SCL   &  0.37 & 0.96\\ \hline
App. No.  & 0.73 & 0.88     \\ \hline
Our Method &  0.89 & 0.86   \\ \hline
\end{tabular}
\caption{Results of Manual Quality Assessment of sampled 120 cases against various approaches, for citation extraction and mapping.}
\label{manual}
\end{table}

\subsection {Dataset splits \& Analysis}
The above construction process resulted in 15,729 ECtHR judgements available in English up until July 2022, each subdivided into the facts section and the reasoning section. For each document, we obtain a valid list of citations extracted from the document. The entire dataset is chronologically split into training (9.7k, 1960–2014), development (2.1k, 2015–2017), and test (3.2k, 2018–2022) sets. Table \ref{datasets_prior} provides a comparison of our dataset with prior datasets. We highlight two ways in which our dataset is distinctively more reflective of a realistic scenario, compared to prior datasets. 

Firstly, unlike prior works which use the whole document as the query and only suppress citation markers, our queries only use the content of the facts section. We posit that the reasoning sections of the case are inexistent before the verdict of the case is finalized and only the facts of the case are available as input to retrieve relevant documents. In contrast, for candidate prior documents, we use both the facts and the reasoning sections (and also compare retrieval performance on each section individually). The stark difference between average number of tokens in query and candidate documents reflects this design choice in our case. 

Secondly, the pool of candidate documents in prior datasets does not reflect a realistic scenario as they are constructed artificially by selecting relevant documents \footnote{While COLIEE uses all the documents cited in the query as relevant, IRLeD samples 5 relevant documents per query.} and randomly sampling irrelevant documents to make a static candidate pool for all the queries. In contrast, we use the entire case-law of ECtHR as the candidate document pool to reflect the scale of a realistic retrieval scenario. Furthermore, our candidate pool is dynamic because for each query, we exclude from the candidate pool those cases dated after the query case.

\input{text/dataset_comparision}

\begin{figure*}
    \centering
    \begin{subfigure}{0.4\textwidth}
    \includegraphics[width=\linewidth]{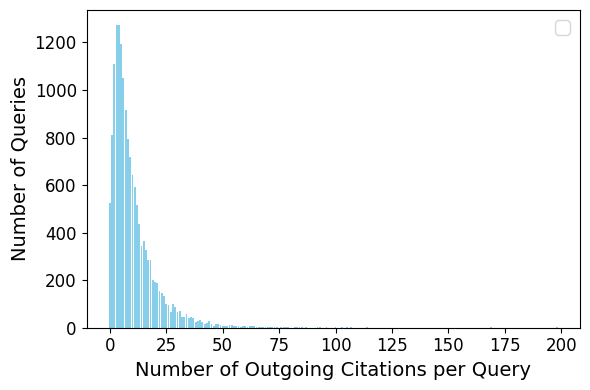}
    \caption{Distribution of Outgoing citations per query document.}
    \label{out_cit}
    \end{subfigure}
    \hfill 
    \begin{subfigure}{0.4\textwidth}
         \includegraphics[width=\linewidth]{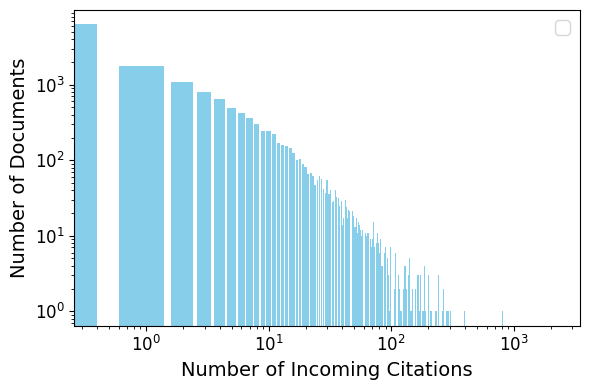}
    \caption{Distribution of Documents based on incoming citations (in log scale)}
    \label{inc_cit}
    \end{subfigure}
    \caption{Analysis of number of Citations/relevant documents.}
    \label{analysis-dataset}
\end{figure*}


In Fig. \ref{out_cit}, we present the distribution of outgoing citations for all documents. Meanwhile, Fig. \ref{inc_cit} depicts the distribution of incoming citations, reflecting how often each document has been referenced by other cases. Roughly 6,000 cases have not received any citations. Notably, about 1,500 of these cases are from the most recent five years, representing over 50\% of cases in this period. This is expected since recent cases typically take time to gain recognition in the legal community. On the other hand, approximately 1,700 cases have been cited exactly once, while the rest have received multiple citations. The frequency of incoming citations serves as an indicator of a case's influence and relevance within the legal domain. Cases that have been cited multiple times likely contain valuable legal arguments, reasoning, or precedents widely recognized and referenced by other cases. Therefore, considering citation frequency as a signal can be beneficial for modeling relevance in prior case retrieval and should be considered in future efforts.

%% file: text/dataset_comparision.tex
\begin{table*}[]
\scalebox{0.97}{
\begin{tabular}{|l|ccc|c|cc|}
\hline
Dataset                       & \multicolumn{3}{c|}{ECtHR-PCR}                                                               & IRLeD   & \multicolumn{2}{c|}{COLIEE}            \\ \hline
Split                         & \multicolumn{1}{c|}{Train}                        & \multicolumn{1}{c|}{Valid}    & Test     & Test    & \multicolumn{1}{c|}{Train}   & Test    \\ \hline
\#Queries                     & \multicolumn{1}{c|}{9787} & \multicolumn{1}{c|}{2186}     & 3231     & 200     & \multicolumn{1}{c|}{898}     & 300     \\ \hline
Avg. \#Candidates per query   & \multicolumn{1}{c|}{5283.22}                      & \multicolumn{1}{c|}{11374.96} & 14102.01 & 2000    & \multicolumn{1}{c|}{4415}    & 1564    \\ \hline
Avg. \#relevant Doc per query & \multicolumn{1}{c|}{9.61}                         & \multicolumn{1}{c|}{12.97}    & 12       & 5       & \multicolumn{1}{c|}{4.68}    & -       \\ \hline
Avg. \#words in query         & \multicolumn{1}{c|}{1706.39}                      & \multicolumn{1}{c|}{1765.11}  & 1743.57  & 7883.41 & \multicolumn{1}{c|}{4628.42} & 5327.08 \\ \hline
Avg. \#words in candidate     & \multicolumn{1}{c|}{5530.43}                      & \multicolumn{1}{c|}{6075.86}  & 5887.91  & 7377.77 & \multicolumn{1}{c|}{4777.98} & 4976.06 \\ \hline
\end{tabular}}
\caption{Statistics of ECtHR-PCR dataset in comparison to other prior PCR datsets.}
\label{datasets_prior}
\end{table*}

%% file: text/models.tex
\textbf{Task Defintion:} The aim of PCR is to learn a retrieval function $R : (q, D) \rightarrow D_r$ that takes as input a query case $q$ (consisting of facts $q_f$ and timestamp $q_t$ ) 
and a corpus of candidate case documents $D$ prior to the case $q$ i.e 
$D = \{d_1, d_2, \ldots, d_{|D|}\}$ (Document $d_i$ consists of facts $d_{if}$, reasoning $d_{il}$ and is associated with a timestamp $d_{it}$ and $\forall_{i \in \{1,2,\ldots\ |D|\}} d_{it} < q_t $) 
and returns a ranked list of candidate case documents $D_r$ based on their relevance score with the query $q$. 

\subsection{Lexical Models}
\noindent \textbf{BM25} \cite{robertson1995okapi} is a bag-of-words based method which computes the relevance score based on the query terms appearing in document.

\subsection{Dense Models}
Lexical approaches rely on bag-of-words matching and face the lexical gap issue, whereby they consider only documents with the same keywords as the query to be relevant. Recent studies \cite{karpukhin2020dense,xiongapproximate} have explored dense-based architectures which can effectively capture semantic relationships between queries and documents by matching them in a continuous representation space learned via deep neural networks. 
 
One of the common approach of dense models, employs a dual-encoder architecture, consisting of a query encoder $E_q$ and a document encoder $E_d$ which transform query q and document d into m-dimensional dense vectors $h_q$ and $h_d$ respectively. Then, the relevance score of d with respect to q can be computed using the dot product.
\noindent To capture the similarity between queries and documents, complex interaction mechanisms like cross-attention can be used. But it is important for the similarity function to be decomposable, enabling pre-computation of document representations for scalability, especially with large document collections. Post training, the document encoder is applied to all documents offline and their final representations are indexed using a datastore. At the run time, when given a query $q$, we obtain its embedding $h_q = E_q(q)$ and retrieve the top k documents with embeddings closest to $h_q$ from the datastore using maximum-inner product search (MIPS). We use FAISS, an efficient open-source library for building datastore data stores of dense vectors which can be queried using similarity search. \cite{johnson2019billion}.
\newline

\noindent \textbf{Encoders:} 
Dense retrieval methods often employ pre-trained language models like BERT \cite{devlin2018bert} and RoBERTa \cite{liu2019roberta} as backbone encoders. However, in our case, both queries and documents are lengthy, exceeding the input length limitation of 512 tokens. Hence, we use a BERT variant of Hierarchical Attention Networks \cite{yang2016hierarchical} as our encoder, adopted from previous works \cite{santosh2022deconfounding}. To handle long input texts, we employ a greedy sentence packing strategy. This strategy involves packing as many sentences as possible into one packet until it reaches the maximum length limit (512 tokens for BERT). Each packet is then encoded using LegalBERT \cite{chalkidis2020legal} to obtain token-level representations $z_i = \{z_{i1}, z_{i2}, \ldots, z_{in}\}$. A token attention layer is utilized to aggregate the token representations into packet vectors as follows: 
\begin{equation}
    u_{it} = \tanh(W_w z_{it} + b_w )  
\end{equation}
\begin{equation}
    \alpha_{it} = \frac{\exp(u_{it}u_w)}{\sum_t \exp(u_{it}u_w)}  
~~\&~~
    f_i = \sum_{t=1}^n \alpha_{it}z_{it}
\label{att}
\end{equation}
where $W_w$, $b_w$ and $u_w$ are trainable parameters and $\alpha_{it}$ represents the importance of $t^\text{th}$ token in the $i^\text{th}$ packet. 
These packet vectors are processed using a GRU encoder to obtain contextual representations. Finally, these are aggregated using a sentence attention layer (similar to Eq. \ref{att}) to obtain the final dense representation of the input. 
We explore two variants of encoders. (i) Uniencoder \cite{xiong2021approximate}, which uses the same query and document encoder in a shared vector space, and (ii) biencoder \cite{karpukhin2020dense}, which uses independent query and document encoder in different embedding spaces.
\newline

\noindent \textbf{Training:} The training objective of the dense text retrieval task is to pull the representations of the query $q$ and relevant documents $D^+$ together, while pushing away irrelevant ones $D^- = D \textbackslash D^+$. We optimize the negative log likelihood of the relevant documents against irrelevant ones as follows:
\begin{equation}
    L(q, D^+, D^-) = -\log \frac{e^ {rel(q,D^+)}}{\sum_{d_k \in D} e^{rel(q ,d_k)}}
\end{equation}
\begin{equation}
     rel(q, d) = E_q(q)^TE_d (d)
\end{equation}
\newline

\noindent \textbf{Negative sampling:} To train a dense retrieval model effectively, proper negative document sampling is crucial, considering the large number of negative documents per query. We explore three different negative sampling strategies. \footnote{We obey the timestamp constraint even with all the negative sampling strategies during training i.e timestamp of query is always greater than timestamp of negative documents.}
\newline

\noindent \textbf{Random} \cite{karpukhin2020dense}: any random documents from the candidate pool which are not relevant to the query.


\noindent \textbf{BM25 + Random} \cite{karpukhin2020dense}:  combines negatives from BM25 and random sampling.
\vspace{1mm}

\noindent \textbf{ANCE} \cite{xiong2021approximate}:  Approximate Nearest Neighbor Negative Contrastive Learning (ANCE)  suggests that random negatives are uninformative, and BM25 negatives overfit the model to lexical signals. ANCE proposes selecting top-k hard negatives ranked by the dense retrieval model itself. Negatives are asynchronously updated after each epoch using the trained model.

%% file: text/experiments.tex
\input{text/results_table}

\subsection{Evaluation Metrics}

We evaluate the performance using Recall@k and Mean Average Precision (MAP). Recall@k measures the proportion of relevant documents ranked in the top k candidates, with k values of 50, 100, 500, and 1000. We report the average Recall@k across all instances. MAP calculates the mean of the Average Precision scores for each instance, where Average Precision is the average of Precision@k scores for every rank position of each relevant document. Precision@k denotes the proportion of relevant documents in top k candidates. Higher scores indicate better performance for Recall@k and MAP. Additionally, we report Mean Rank and Median Rank, which are the averages of the median and mean ranks of relevant documents for each instance. Lower values are preferred for both of these metrics

\subsection{Implementation Details}
\label{imp_details}
\textbf{BM25:} The hyperparameters $K_1$ and $b$ are sweeped in the range[0, 3] and [0, 1]  respectively, to pick the best value based on the validation set. 
\newline

\noindent \textbf{Dense Models:} For all of our dense models, we use LegalBERT \cite{chalkidis2020legal} as the backbone, which produces a word embedding size of 768. Our word level attention context vector size is 300. The sentence level GRU encoder dimension is 200, thus giving a bidirectional embedding of size 400, and a sentence level attention vector dimension of 200. We use 4 queries in a batch with 1 positive and 7 negatives per query, resulting in 32 cases per batch. The model is optimized end- to-end using Adam \cite{kingma2014adam}.  We determine the best learning rate using a grid search on the validation set and use early stopping based on the development set and the Recall@1000 score.

\subsection{Results}
We report the results on the test set of the ECtHR- PCR dataset in Tables \ref{results-unibi}, \ref{tab-neg} and \ref{halsbury}. In both these tables, note that we use only the facts for query case and both the facts and the law section for documents. 
\vspace{2mm}

\noindent \textbf{BM25 vs Dense Models:} 
BM25 performs competitively with dense models and even better than DR-Uniencoder across most of the metrics. This can be attributed to the high lexical overlap and repeated usage of legal tests and concepts in legal cases as effectively captured by BM25. While dense modeling excels at capturing semantic relationships and contextual information, it seems to fail to capture the lexical overlap of the legal documents, as evidenced by the DR-Uniencoder's lower performance. BM25 proves particularly more effective at lower k values, 
underlining how high lexical overlap provides a better signal of relevance.
\newline

\noindent \textbf{Uniencoder vs Biencoder:} 
We observe that the biencoder model significantly outperforms the uniencoder model across all metrics, surpassing even BM25 at higher k values. This can be attributed to the differing semantics between queries and documents, as queries contain only the factual statements of the cases while documents contain both the factual statements and the reasoning section. This necessitates two separate encoders to effectively capture these distinct aspects. 
\newline

\begin{figure}
    \centering
    \includegraphics[width = 0.45\textwidth]{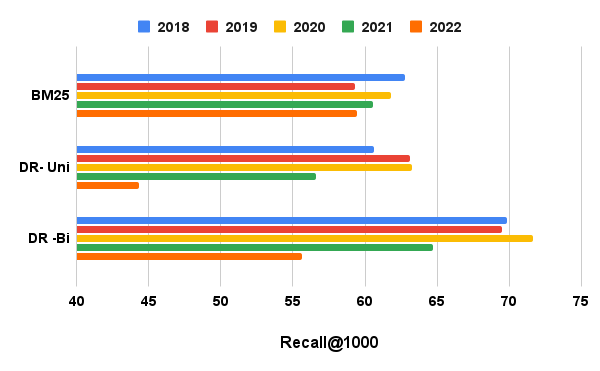}
    \caption{Comparison of BM25 and DR-(Uni/Bi) trained with random sampling over different years in the test set.}
    \label{temporal-shift}
\end{figure}

\input{text/results_table2}

\noindent \textbf{Impact of negative sampling: } 
We use the DR-Biencoder architecture and vary different negative sampling strategies and report their performance in Table \ref{tab-neg}. DR-Rand model is significantly better than the DR-BM25+Rand and the DPR-ANCE model in all the evaluation metrics. This reveals that training on difficulty-based hard negatives yielded lower performance compared to training on randomly selected negatives, which is surprising considering both hard-negative strategies were effective in traditional IR tasks. It is possible that these hard-negative selection strategies, by selecting negatives which are lexically (BM25) or semantically (ANCE) similar to the query, are actually selecting relevant documents that simply have not been cited (false negatives), a fact that would be less problematic in the traditional question-based querying of standard information retrieval research. Comparing  difficulty-based strategies, ANCE sampling performs better than BM25+Rand,  highlighting that focusing on distinguishing lexically similar candidates through BM25 sampling, results in over specialization that neglects broader variants of negatives in the entire corpus.

\subsection{Temporal Degradation}
We assess how DR-Uni and -Bi models perform over time and we present the Recall@1000 scores for BM25 and DR-Rand models for both architectures over the test set period (2018- 22), separately for each year in Fig. \ref{temporal-shift}. Both the dense models exhibit a more pronounced decline compared to BM25 in recent years due to temporal distributional shift of ML models - difference between the distribution of the training and the test data caused by the addition of new case documents to the candidate pool over time. In the earlier years of the test set (2018-2020), most citations come from cases seen during training, but in later years citations point towards unseen cases which are added subsequently (directly in validation set or even early years of test set). To address this temporal drift, future research should focus on continual updating of dense encoders with arrival of new candidate documents and enhancing model robustness to temporal shifts in the data, enabling better adaptation to evolving legal landscapes.

\subsection{ECtHR: Halsbury's or Goodhart's view ?}
We quantitatively assess the notion of relevance as per Halsbury’s and Goodhart’s views in practice, by using only the reasoning and the facts section for the documents in candidate pool, respectively. From Table \ref{halsbury}, we observe that using law section of the document alone turns more effective than using the facts section alone, as witnessed across the three model variants. This finding provides more evidence supporting Halsbury’s view in the ECtHR domain compared to Goodhart’s view, which aligns with the findings of \citealt{valvoda2021precedent}. 

Comparing Table \ref{halsbury} to Table \ref{results-unibi}, we observe that using the law section alone proves more effective than using the entire document. This can be attributed to the fact that the law section attempts to provide arguments on why a certain article is violated by grounding them in the particular circumstances of the case. In other words, the important premises of the facts section are discussed in the law section, allowing the model to directly focus on the relevant factual information presented in the law section. Further, adding facts along with the reasoning section tends to shift the model's focus with unnecessary additional information, leading to a drop in performance. Further investigation is required to analyze these conjectures in relation to each of the alleged articles present in each case.

\subsection{Discussion}
Overall, we notice a low score across the board and highlight some challenges and directions to be pursued to build effective prior case retrieval systems. One limitation of our proposed baselines is their focus solely on the text content, neglecting the potential insights from the citation network 
with its interconnected relationships 
providing a rich global view of case law. 

Another significant challenge arises with dynamic evolution of law with time incorporating changes in norms, societal attitudes, and values and time-dependent nature of precedents, as they can be overruled, resulting in the expansion or contraction of the law. Notably, the ECHR convention was intentionally drafted in an abstract manner to allow for interpretation and to encompass a wide range of situations, distinguishing it from more specific national legal codes. Exploring methods to capture the temporal nature of precedents would be an interesting direction. 


Furthermore, in order to achieve a comprehensive understanding of relevance in prior case retrieval, it is crucial for an ideal PCR model to not only comprehend the case facts but also deduce the reasoning process that can be applied to the query facts. The model would be then able to identify prior cases that support the reasoning process. This can be facilitated by allowing the model to generate the reasoning process initially and then retrieve the document that substantiates it. By incorporating such mechanisms, the model enhances its explainability by providing a rationale for the relevance of a specific candidate document

Exploration of above discussed strategies such as incorporating various nuances of citation modeling for better relevance understanding, has been hindered due to limitations in prior datasets. These limitations include the lack of a complete citation network, timestamp information of documents, fragmented splits of candidate document pool and artificially constructed snapshots of the data by randomly selecting a few relevant and irrelevant cases for the candidate document pool. However, our ECtHR-PCR dataset overcomes these limitations by providing the entire case law of the European Court of Human Rights since its inception. With this comprehensive dataset, researchers now have the opportunity to explore realistic relevance modeling that emulates how legal experts select prior cases for citation, enabling a more accurate representation of the complexities involved.

%% file: text/results_table.tex
\begin{table*}[]
\centering
\begin{tabular}{|l|cccc|c|c|c|}
\hline
\textbf{}             & \multicolumn{4}{c|}{\textbf{Recall@k ($\uparrow$)}}                                                                                   & \textbf{}    & \multirow{2}{*}{\textbf{\begin{tabular}[c]{@{}c@{}}Mean\\ Rank ($\downarrow$) \end{tabular}}} & \multirow{2}{*}{\textbf{\begin{tabular}[c]{@{}c@{}}Median\\ Rank ($\downarrow$) \end{tabular}}} \\ \cline{1-6}
\textbf{Model}        & \multicolumn{1}{c|}{\textbf{50}} & \multicolumn{1}{c|}{\textbf{100}} & \multicolumn{1}{c|}{\textbf{500}} & \textbf{1000} & \textbf{MAP ($\uparrow$)}  &                                                                               &                                                                                 \\ \hline
BM25                  & \multicolumn{1}{c|}{22.14}       & \multicolumn{1}{c|}{27.82}        & \multicolumn{1}{c|}{47.8}         & 60.38         & 9.65         & 1945.73                                                                       & 1218.07                                                                         \\ \hline
DR-Rand-Uniencoder & \multicolumn{1}{c|}{19.33}       & \multicolumn{1}{c|}{26.19}        & \multicolumn{1}{c|}{47.61}        & 58.9          & 7.28         & 1827.08                                                                       & 1388.38                                                                         \\ \hline
DR-Rand-Biencoder   & \multicolumn{1}{c|}{20.36}       & \multicolumn{1}{c|}{29.26}        & \multicolumn{1}{c|}{56.03}        & 67.31         & 6.72         & 1676.55                                                                       & 1387.8                                                                          \\ \hline
\end{tabular}
 \caption{ Retrieval performance on test set using BM25 and dense models based on uniencoder and biencoder architectures, trained with random sampling. 
 $\uparrow$ ($\downarrow$) indicates higher(lower) the value, better the performance. Recall@k and MAP are expressed in percentages.}
\label{results-unibi}
\end{table*}

\begin{table*}[]
\centering
\begin{tabular}{|l|cccc|c|c|c|}
\hline
\multicolumn{1}{|c|}{\textbf{}} & \multicolumn{4}{c|}{\textbf{Recall@k} ($\uparrow$)}                                                                                   & \textbf{}    & \multirow{2}{*}{\textbf{\begin{tabular}[c]{@{}c@{}}Mean\\ Rank ($\downarrow$)\end{tabular}}} & \multirow{2}{*}{\textbf{\begin{tabular}[c]{@{}c@{}}Median\\ Rank ($\downarrow$)\end{tabular}}} \\ \cline{1-6}
\textbf{Model}                  & \multicolumn{1}{c|}{\textbf{50}} & \multicolumn{1}{c|}{\textbf{100}} & \multicolumn{1}{c|}{\textbf{500}} & \textbf{1000} & \textbf{MAP ($\uparrow$)} &                                                                               &                                                                                 \\ \hline
DR-Rand                        & \multicolumn{1}{c|}{20.36}       & \multicolumn{1}{c|}{29.26}        & \multicolumn{1}{c|}{56.03}        & 67.31         & 6.72         & 1676.55                                                                       & 1387.8                                                                          \\ \hline
DR-BM25+Rand                   & \multicolumn{1}{c|}{13.65}       & \multicolumn{1}{c|}{18.8}         & \multicolumn{1}{c|}{38.51}        & 51.72         & 5.35         & 2275.41                                                                       & 1944.51                                                                         \\ \hline
DR-ANCE                        & \multicolumn{1}{c|}{15.38}       & \multicolumn{1}{c|}{22.63}        & \multicolumn{1}{c|}{45.97}        & 57.4          & 4.9          & 2101.06                                                                       & 1703.8                                                                          \\ \hline
\end{tabular}
\caption{Retrieval performance on test set using DR-Biencoder architecture, trained using different negative sampling strategies. $\uparrow$ ($\downarrow$) indicates higher(lower) the value, better the performance. Recall@k and MAP are expressed in percentages. }
\label{tab-neg}
\end{table*}

%% file: text/results_table2.tex
\begin{table*}[]
\centering
\scalebox{0.9}{
\begin{tabular}{|l|c|cccc|c|c|c|}
\hline
\textbf{}             & \multirow{2}{*}{\textbf{\begin{tabular}[c]{@{}c@{}}Document \\ Elements \end{tabular}}} & \multicolumn{4}{c|}{\textbf{Recall@k ($\uparrow$)}}                                                                                   & \textbf{}    & \multirow{2}{*}{\textbf{\begin{tabular}[c]{@{}c@{}}Mean\\ Rank ($\downarrow$)\end{tabular}}} & \multirow{2}{*}{\textbf{\begin{tabular}[c]{@{}c@{}}Median\\ Rank ($\downarrow$)\end{tabular}}} \\ \cline{1-1} \cline{3-7}
\textbf{Model}        &                                                                                                & \multicolumn{1}{c|}{\textbf{50}} & \multicolumn{1}{c|}{\textbf{100}} & \multicolumn{1}{c|}{\textbf{500}} & \textbf{1000} & \textbf{MAP ($\uparrow$)} &                                                                               &                                                                                 \\ \hline
BM25                  & Facts                                                                                          & \multicolumn{1}{c|}{19.54}       & \multicolumn{1}{c|}{24.81}        & \multicolumn{1}{c|}{41.98}        & 52.55         & 8.24         & 2577.65                                                                       & 1802.26                                                                         \\ 
DR-Rand - UniEncoder & Facts                                                                                          & \multicolumn{1}{c|}{18.89}       & \multicolumn{1}{c|}{25.21}        & \multicolumn{1}{c|}{44.33}        & 54.07         & 7.23         & 2205.01                                                                       & 1688.33                                                                         \\ 
DR-Rand- BiEncoder   & Facts                                                                                          & \multicolumn{1}{c|}{17.04}       & \multicolumn{1}{c|}{24.65}        & \multicolumn{1}{c|}{49.76}        & 60.8          & 5.41         & 2225.31                                                                       & 1797.27                                                                         \\ \hline
BM25                  & Law                                                                                            & \multicolumn{1}{c|}{22.72}       & \multicolumn{1}{c|}{28.67}        & \multicolumn{1}{c|}{52.25}        & 62.47         & 10.26        & 1824.59                                                                       & 1053.12                                                                         \\ 
DR-Rand - UniEncoder & Law                                                                                            & \multicolumn{1}{c|}{19.22}       & \multicolumn{1}{c|}{26.62}        & \multicolumn{1}{c|}{50.28}        & 62.38         & 7.31         & 1503.02                                                                       & 1034                                                                            \\ 
DR-Rand- BiEncoder   & Law                                                                                            & \multicolumn{1}{c|}{23.72}       & \multicolumn{1}{c|}{32.23}        & \multicolumn{1}{c|}{56.08}        & 66.85         & 7.87         & 1572.15                                                                       & 1217.87                                                                         \\ \hline
\end{tabular}}
\caption{Retrieval performance on test set using BM25 and dense models with random sampling. Documents elements section indicate which section of document is specifically used, while results in Table \ref{results-unibi}, \ref{tab-neg} use both the facts and the law section of the document. Note, we only use facts section for the query.}
\label{halsbury}
\end{table*}

%% file: text/conclusion.tex
In this work, we present the ECtHR-PCR, prior case retrieval dataset for jurisdiction of European Court of Human Rights. We benchmark various retrieval baselines, including both lexical-based and dense retrieval models and observe that difficulty-based negative sampling underperforms random negative sampling, highlighting the need to design domain-specific difficult criterion to train effective dense retrieval systems. We demonstrate how neural dense models degrade with time, while BM25 is temporally robust, highlighting the need of developing retrieval models which can handle temporal distributional shifts. Furthermore, we empirically examine the contested Halsbury's and Goodhart's view on what constitutes a ratio based on relevance signal and witness a prevailing trend of Halsbury's view that reasoning and arguments hold more weight in determining relevance.  
In future, we intend to develop retrieval models that incorporate explicit modeling of the citation network, taking into account the specific characteristics of citation behavior in case law documents. 
We hope that this data resource will be useful to the research community working on problems in the space of legal language processing and information retrieval.

%% file: text/limitations.tex

It is important to acknowledge that the ECtHR-PCR dataset created in this study may exhibit a linguistic bias, as French is the other official language of ECtHR along with English and hence some documents in these English judgements may cite non-English ones which have been filtered out. Some of them might indeed contain relevant information needed but not considered in our dataset version. Future efforts should be taken to improve both the recall and precision of our dataset. For instance, we could develop more heuristics to decode the citation string format and map them back to the document to add the relevant documents or we could develop further tigh

Dealing with legal corpora presents technical challenges, particularly due to the lengthy nature of legal texts. To address this, we employ hierarchical models, which have inherent limitations in terms of the direct interaction between tokens across long distances. While this restriction is still an underexplored aspect of hierarchical models, recent work such as \cite{dai2022revisiting} provides preliminary insights in this direction. Furthermore, to optimize computational resources, we freeze the weights in the LegalBERT sentence encoder. It is important to note that there are other limitations in our modeling approach, which have been discussed earlier.

In terms of evaluating the quality of the retrieved relevant documents, we plan to conduct human evaluation in the future to assess the utility of our system, which would be challenging given the length and complexity of the legal text involved which requires annotators with a deep understanding of ECHR jurisprudence to understand and evaluate the notion of relevance. It is worth noting that the models explored in this work primarily act as pre-fetchers, prioritizing recall to ensure that all relevant cases are retrieved. However, in practice, end-users expect a high-quality retrieval system that precisely identifies a smaller number of relevant cases. This requires an additional re-ranker step in the retrieval pipeline to optimize the precision of the ranked list. In this study, we focus on the first step of the retrieval pipeline, leaving the development of a re-ranker component for future work.

%% file: text/ethics.tex
We compiled this dataset by sourcing ECtHR decisions from the publicly accessible HUDOC website. These decisions, although not anonymized, include the real names of individuals involved. However, our work does not engage with the data in a way that we consider harmful beyond this availability. Rather, we believe development of effective PCR would further assist legal professionals, to deal with the increasing volume of cases. We employ pre-trained language models and do not train them from scratch, thus inheriting the biases they may have acquired from their training corpus. We recognize the need for thorough analysis and mitigation of any biases that may arise and it is is crucial to ensure that the PCR systems we develop are fair, unbiased, and uphold principles of equality and justice.